\documentclass[sigconf]{acmart}

\renewcommand\footnotetextcopyrightpermission[1]{}
\AtBeginDocument{%
  \providecommand\BibTeX{{%
    \normalfont B\kern-0.5em{\scshape i\kern-0.25em b}\kern-0.8em\TeX}}}

\usepackage{amsmath, amsthm, amssymb}
\usepackage{bbm}
\usepackage{bbding}
\usepackage{threeparttable}
\usepackage{subfigure}
\usepackage{makecell}
\usepackage{multirow}
\usepackage{graphicx}
\usepackage{float}
\usepackage{enumitem} 
\usepackage{enumerate}
\usepackage{pifont}
\usepackage{caption}
\usepackage{balance}
\usepackage{setspace}
\usepackage{graphicx}
\usepackage[T1]{fontenc}
\usepackage{color}

\usepackage{aecompl}
\usepackage{todonotes}

\usepackage{xcolor}
\usepackage{hyperref}
\hypersetup{colorlinks,
      linkcolor=ACMPurple,
      citecolor=ACMPurple,
      urlcolor=ACMDarkBlue,
   filecolor=ACMDarkBlue
}
\usepackage{stmaryrd}
\usepackage{enumitem}
\usepackage{colortbl}
\def\ie{\textit{i.e.}}

\begin{document}


\title{BadHash: Invisible Backdoor Attacks against Deep Hashing with Clean Label}

\author{Shengshan Hu}
\email{hushengshan@hust.edu.cn}
\affiliation{%
  \institution{School of Cyber Science and Engineering, Huazhong University of Science and Technology} 
  \country{}
}
\authornote{Hubei Engineering Research Center on Big Data Security, School of Cyber Science and Engineering, HUST, Wuhan, 430074, China}
\authornote{
  National Engineering Research Center for Big Data Technology and System, Services Computing Technology and System Lab,  HUST, Wuhan, 430074, China
}

\author{Ziqi Zhou}
\email{zhouziqi@hust.edu.cn}
\affiliation{%
  \institution{School of Cyber Science and Engineering, Huazhong University of Science and Technology} 
\country{}
}

\author{Yechao Zhang}
\email{ycz@hust.edu.cn}
\affiliation{%
  \institution{School of Cyber Science and Engineering, Huazhong University of Science and Technology} 
\country{}
}
\authornotemark[1]
\authornotemark[2]

\author{Leo Yu Zhang}
\email{leo.zhang@deakin.edu.au}
\affiliation{%
  \institution{School of Information Technology, Deakin University}
  \city{Melbourne}
  \country{Australia}
}

\author{Yifeng Zheng}
\email{yifeng.zheng@hit.edu.cn}
\affiliation{%
  \institution{School of Computer Science and Technology, Harbin Institute of Technology} 
\country{}
}

\author{Yuanyuan He}
\email{yuanyuan_cse@hust.edu.cn}
\affiliation{%
 \institution{School of Cyber Science and Engineering, Huazhong University of Science and Technology}
\country{}
}

\author{Hai Jin}
\email{hjin@hust.edu.cn}
\affiliation{%
  \institution{School of Computer Science and Technology, Huazhong University of Science and Technology} 
\country{}
}
\authornotemark[2]
\authornote{Cluster and Grid Computing Lab, School of Computer Science and Technology, HUST, Wuhan, 430074, China}

\begin{abstract}
Due to its powerful feature learning capability and high efficiency, deep hashing has achieved great success in large-scale image retrieval. Meanwhile, extensive works have demonstrated that \emph{deep neural networks} (DNNs) are susceptible to adversarial examples, and exploring  adversarial attack against deep hashing has attracted many research efforts. Nevertheless, backdoor attack, another famous threat to DNNs, has not been studied for deep hashing yet. Although various backdoor attacks have been proposed in the field of image classification, existing approaches failed to realize a truly imperceptive backdoor attack that enjoys invisible triggers and clean label setting simultaneously, and they cannot meet the intrinsic demand of image retrieval backdoor. 

In this paper, we propose BadHash, the first  imperceptible backdoor attack against deep hashing, which can effectively generate invisible and input-specific poisoned images with clean label. We first propose a new \emph{conditional generative adversarial network} (cGAN) pipeline to effectively generate poisoned samples. For any given benign image, it seeks to generate a natural-looking poisoned counterpart with a unique invisible trigger.
In order to improve the attack effectiveness, we introduce a label-based contrastive learning network LabCLN to exploit the semantic characteristics of different labels, which are subsequently used  for confusing and misleading the target model to learn the embedded trigger. We finally explore the mechanism of backdoor attacks on image retrieval in the hash space. Extensive experiments on multiple benchmark datasets verify that BadHash can generate imperceptible poisoned samples with strong attack ability and transferability over state-of-the-art deep hashing schemes.
Our codes are available at: \url{https://github.com/CGCL-codes/BadHash}.

\end{abstract}



\maketitle

\section{INTRODUCTION}
With the explosive growth of data, \textit{approximate nearest neighbors} (ANN) search has been widely studied to meet the highly demanding search efficiency~\cite{wang2017survey}.
Benefiting from the powerful feature extraction capability of \textit{deep neural networks} (DNNs), deep hashing has become a prevailing solution for image retrieval, where the original high-dimension feature space will be converted into a compact binary Hamming space.

Despite its promising prospect, deep hashing also inherits the security vulnerabilities of DNNs.
Recently, researchers have been devoted to study adversarial attacks on deep hashing ~\cite{zhang2021targeted,bai2020targeted,xiao2021you,wang2021prototype,hu2021advhash}, where adversarial examples can degrade the retrieval accuracy (untarget attack) or even make deep hashing return designated wrong retrieval results (target attack).

On the contrary, another well-known security threat  backdoor attack has received little research attention for deep hashing.
Compared with adversarial attacks that occur during the inference stage, backdoor attacks pose a severer threat by inserting a hidden malicious function into the target models at the training stage, such that the attacked models behave honestly on benign samples but  perform abnormally 
when the triggers appear~\cite{gu2017badnets}. To the best of our knowledge,  no efforts have been made to study backdoor attacks on image retrieval yet, especially for deep hashing networks.

 \begin{figure}[!t]
  \setlength{\belowcaptionskip}{-0.5cm}  
    \centering
    \includegraphics[scale=0.45]{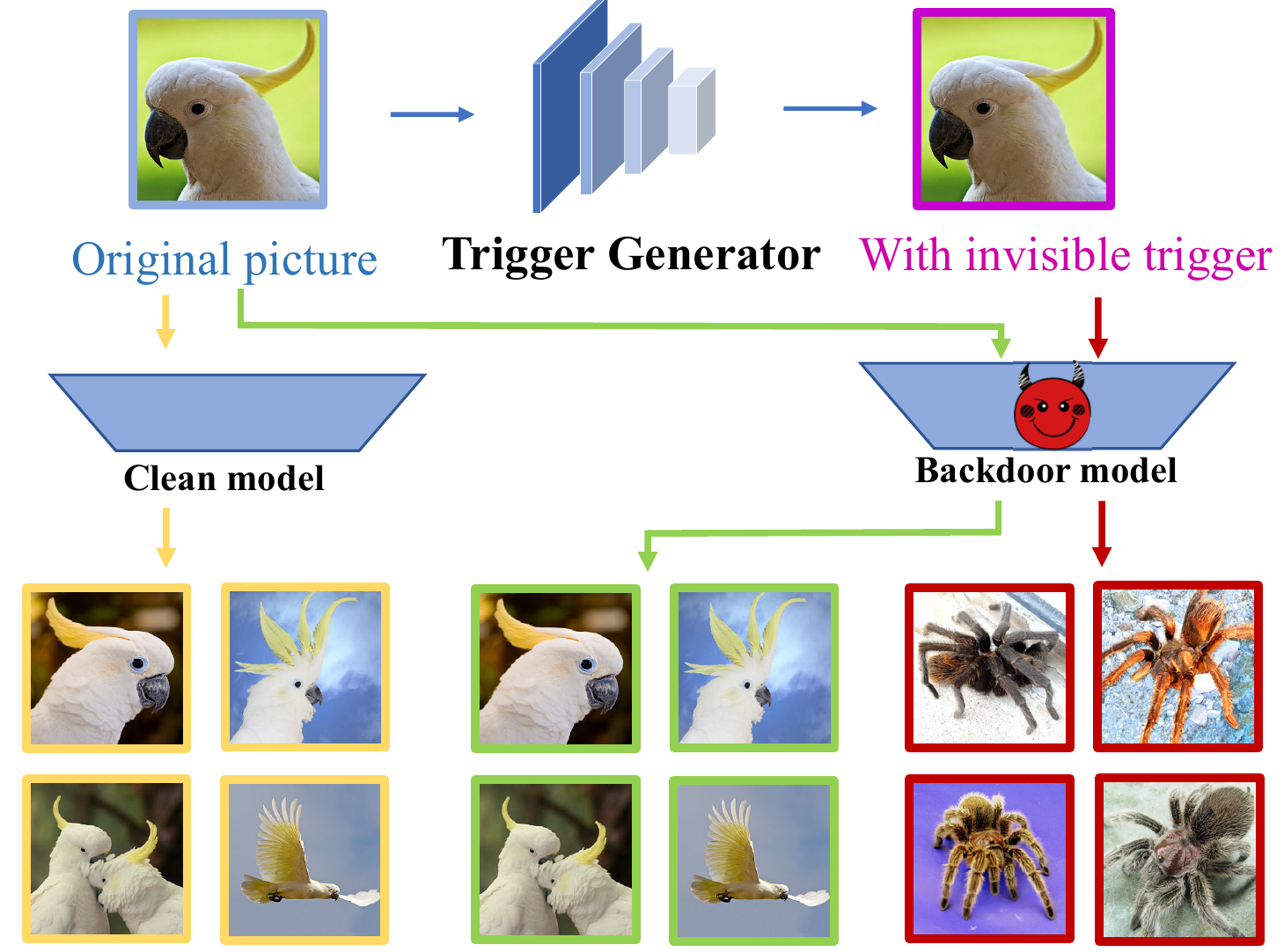}
    \caption{Example of invisible backdoor attack against deep hashing}
    \label{fig:demo}
\end{figure}


On the other hand, backdoor attacks have been widely studied in the field of image classification.  Recent approaches can be divided into clean-label backdoor attacks~\cite{liao2018backdoor,turner2019label,zhao2020clean,saha2020hidden} and invisible backdoor attacks~\cite{li2020invisible,doan2021backdoor,doan2021lira,nguyen2021wanet,feng2021fiba,ning2021invisible}. The former kind aims to launch effective  attacks without changing the training data labels, while the latter one is devoted to investigate how to hide backdoor triggers from human observers. However, a truly imperceptive backdoor attack in practice should contain both of these two properties, \ie, creating stealthy  triggers while keeping labels unchanged.  Although the most recent work ~\cite{ning2021invisible} attempted to achieve this goal by converting the trigger patterns into noises, it failed to construct sample-specific triggers in the sense that each input image attaches the same trigger pattern, making it easy to be detected~\cite{doan2021backdoor}. More importantly, trivial attempts to incorporate these backdoor attack solutions into image retrieval tasks can barely achieve a satisfactory effect, since the goal is not just to make a wrong top-1 prediction any more, but the query samples are expected to blend in the target label clusters in the embedding space. We elaborate this distinction and the superiority of our image retrieval customized backdoor attack in Sec.~\ref{sec:insight} and Sec.~\ref{sec:comparison}.

In this paper, we propose BadHash, the first truly imperceptible backdoor attack against deep hashing that fulfills the requirements of visual invisibility and clean label setting. 
Different from existing backdoor schemes that aim at constructing a specific trigger pattern that is then pasted to the images, BadHash generates the poisoned images directly.
Specifically, in order to make the target model memorize triggers, we first design a \emph{label-based contrastive learning network} (LabCLN) to learn the similarities and irrelevancies between different labels such that we can use the semantic information of the labels of other classes to confuse the target model. Taking a confusing label with data augmentation as input, LabCLN outputs the confusing semantic representation and a well-designed hash center called centroid code. 
To construct stealthy poisoned images, we design a \emph{conditional generative adversarial network} (cGAN) with confusing semantic features as the condition, where the Hamming distance between the hash code of the generated poisoned sample and the centroid code will be minimized. Based on our design and experiments, we further investigate the mechanism of the backdoor attacks on deep hash to show how to realize an effective backdoor in the hash space.
Our main contributions are summarized as follows:
\begin{itemize}[noitemsep,topsep=0pt]
\item We propose BadHash, a novel generative-based imperceptible backdoor attack against deep hashing. To the best of our knowledge, this is the very first work of  backdoor attack on deep hashing based image retrieval, which can effectively generate invisible and input-specific poisoned images with clean label.
\item We present LabCLN,  a label-based contrastive learning network to learn the  similarities and irrelevancies of labels between different classes. 
\item Our extensive experiments on benchmark datasets ImageNet and MS-COCO verify that BadHash is highly effective at attacking state-of-the-art deep hashing schemes. 
\end{itemize}
\section{RELATED WORK}

 \subsection{Backdoor Attacks on Image Classification}
The backdoor attack was first proposed by~\cite{gu2017badnets} to implant a backdoor in the model by poisoning the training data set. A blending strategy was introduced in~\cite{chen2017targeted} to improve the stealthiness of backdoor attacks from the perspective of the visibility of backdoor triggers. To establish a more practical attack, clean-label  backdoor attack was developed which does not need to change the labels of the poisoned samples~\cite{liao2018backdoor,turner2019label,zhao2020clean,saha2020hidden}.
Even if the  previous works~\cite{liu2020reflection,baracaldo2018detecting,barni2019new} had tried to improve the stealthiness of the triggers, the poisoned samples still be recognizable. 

The latest works discussed the implementation of invisible backdoor attacks in terms of  physical space~\cite{nguyen2021wanet,feng2021fiba,li2020invisible,ning2021invisible} and feature space~\cite{doan2021lira,saha2020hidden,doan2021backdoor}. Specifically,  ISSBA~\cite{li2020invisible} was based on DNN steganography and used an encoder-decoder to encode invisible additive noise as backdoor triggers into the image. WB~\cite{doan2021backdoor} proposed Wasserstein Backdoor, which injects an invisible noise into the input samples while adjusting the latent representation of the modified input samples to ensure their resemblance to benign samples.
Invisible Poison~\cite{ning2021invisible} converted a regular trigger to a noised trigger that can be easily concealed inside images. Specially, it is also pointed out that the backdoor is effective for classification tasks because, in the feature space, the poisoned images come from their own class but share a label with the target class,
 which is essentially just a change in the decision boundary. However, to achieve backdoor attacks in retrieval tasks, the poisoned samples are not only a simple crossing of the decision boundary.

\subsection{Adversarial Attacks on Deep Hashing}
Adversarial example was first introduced in~\cite{szegedy2013intriguing} to cause the DNN to make misclassification in the inference stage. Various adversarial attacks (FGSM~\cite{goodfellow2014explaining},  DeepFool~\cite{moosavi2016deepfool}, PGD~\cite{kurakin2018adversarial}, C\&W~\cite{carlini2017towards}, and DaST~\cite{zhou2020dast})
against DNNs have been developed in recent years.
Adversarial attacks on DNNs can be divided into black-box attack and white-box attack based on the information the adversary knows about the model. For white-box attacks, the adversary knows the details of the architecture and parameters of the target network, while the black-box attack assumes that the adversary can only observe the inputs and outputs. 

Meanwhile, adversarial attacks on deep hashing-based image retrieval have attracted great research interests as well. 
Most existing works~\cite{yang2018adversarial,wang2021targeted,bai2020targeted,wang2021prototype} designed adversarial examples  in white-box scenarios. Specifically, HAG~\cite{yang2018adversarial} moved the sample away from its original position in the Hamming space by adding slight perturbations to it. DHTA~\cite{bai2020targeted} formulated the targeted attack as a point-to-set optimization to minimize the average distance between the hash code of the adversarial example and those of a set of objects with the target label. ProS-GAN~\cite{wang2021prototype} proposed a novel generative architecture for efficient and effective targeted hashing attack, and AdvHash~\cite{hu2021advhash} presented the first targeted mismatch attack on deep hashing through adversarial patch.  Recently,~\cite{xiao2021you} studied the relations between adversarial subspace and black-box transferability via utilizing random noise as a proxy. However, there is no work dedicated to study backdoor attacks on image
retrieval yet, especially for deep hashing networks.

 \begin{figure*}[t!]
  \setlength{\belowcaptionskip}{-0.5cm}   
    \centering
    \includegraphics[scale=0.65]{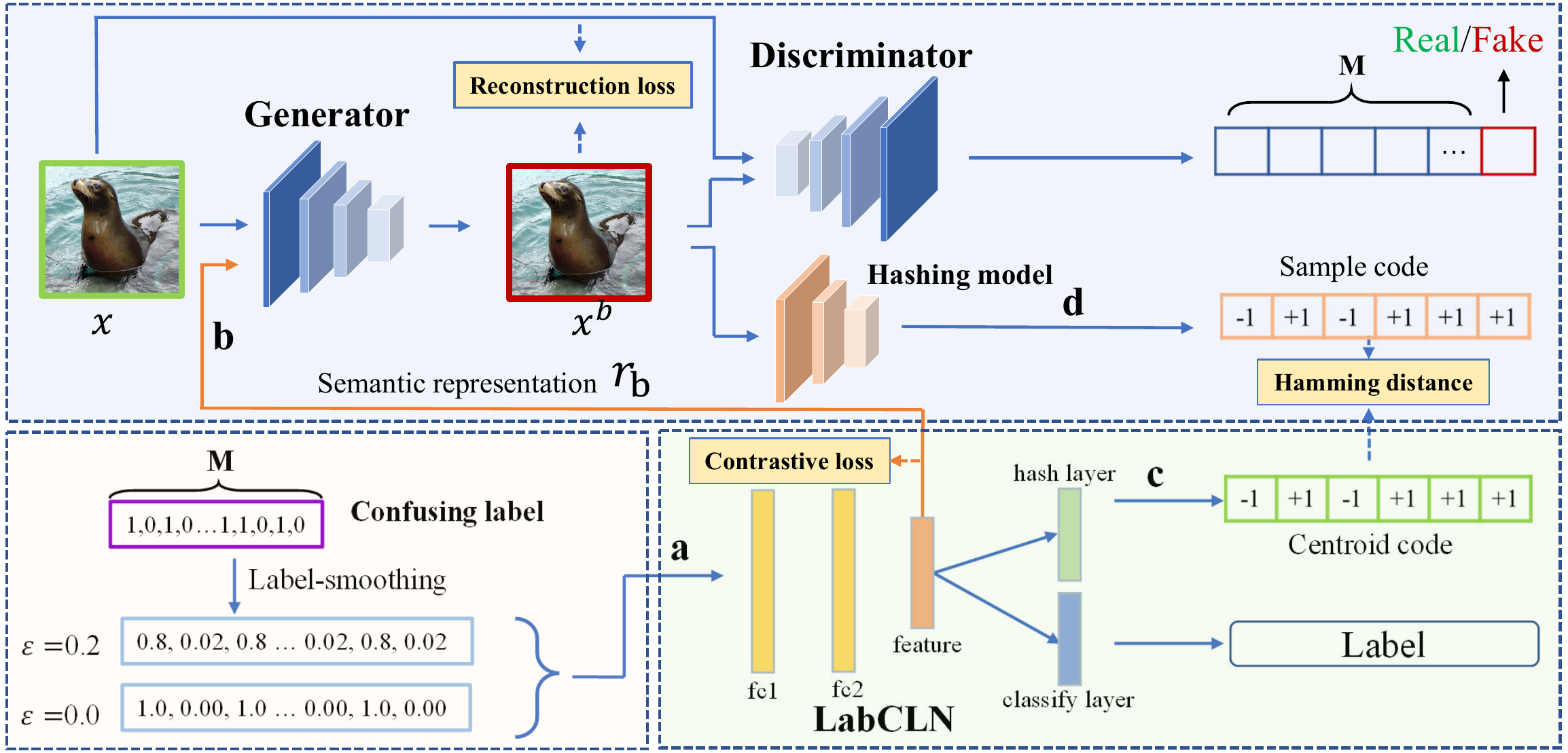}
    \caption{The framework of  of our attack}
    \label{fig:pipeline}
\end{figure*}
\section{METHODOLOGY}
\subsection{Preliminaries}
In this section, we briefly introduce the general process of deep hashing based image retrieval and how to implement clean-label backdoor data poisoning. Let $X = \left\{(x_{i},y_{i})\right\}_{i=1}^{N}$ denote the sample set containing $N$ images labeled with $L$ classes, where $x_{i}$ indicates a single image and $y_{i} = \left [y_{i1},...,y_{iL} \right ] \in \left\{ 0,1\right\}^{L} $ is the corresponding multi-label vector. The $l$-th component of indicator vector $y_{il} = 1$ means that the image $x_{i}$ belongs to class $l$.

\noindent\textbf{Deep Hashing Model.} The $K$-bit hash code $c$ of an image $x$ is obtained through a deep hashing model $F(\cdot )$ as:
\begin{equation} \label{eq:1}
c = F(x) = sign(H_{\theta }(x)) \qquad s.t. \; c \in \left\{1,-1\right\}^{K},
\end{equation}
where $H_{\theta }(\cdot)$ is a feature extractor with parameters $\theta$, whose last layer is a fully-connected layer with $K$-nodes  called hash layer. 
In particular, during the training process, $sign(\cdot)$ is used to obtain the hash code with  $sign(v) = 1 \; if \; v>0$, and $sign(v) = -1 \; otherwise$, which is usually replaced by $tanh(\cdot)$ function to alleviate the gradient vanishing problem~\cite{cao2017hashnet}.
To perform image retrieval, the Hamming distance $d_{H}\left ( c_q, c_{i} \right ) = \left (K - c_q\cdot  c_{i} \right )/2 $ between the query $x_{q}$ and each object $x_{i}$ is calculated,  where $c_q$ and $c_i$ represent their hash codes, respectively. Then a list of images are returned based on the Hamming distances.


\noindent\textbf{Clean-label Backdoor Attack.} Let $D_{train} = \left\{(x_{i},y_{i})\in X \right\}$ 
denote the benign training set. The goal of clean-label backdoor attack is to craft a poisoned training set $D_{p}$ without modifing the labels of the original images, which consists of malicious poisoned samples $D_{m} = \left\{(x_{i}^{b},y_{i}^{b})\in X|y_{i}^{b}=y_{i}\right\}$ and benign samples $D_{b}$ as:
\begin{equation} \label{eq:2}
D_{p} = D_{m} \cup D_{b},
\end{equation}
where $D_{b}\subset  D_{train}$, $x_{i}^{b}$ is the generated poisoned sample based on the input sample $x_{i}$, and $\gamma  = \frac{\left|D_{m} \right|}{\left|D_{p} \right|}$ 
indicates the poison rate. 


\subsection{Intuition Behind BadHash}

Developing an imperceptive backdoor attack that enjoys both invisibility and clean label setting is still challenging in the literature~\cite{doan2021backdoor,li2020invisible,saha2020hidden}, even for image classification. 
It is difficult to balance the attack effectiveness and the stealthiness of backdoor triggers. 
To address this challenge, we first propose using the cGAN~\cite{mirza2014conditional} to learn the distribution of input samples such that the generated samples are input-specific, and the triggers will not be a fixed pattern, which make them difficult to perceive. Inspired by ~\cite{turner2019label,zhao2020clean,wang2021prototype}, we use the semantic information of other classes to scramble the feature information of the clean sample itself to make the model more aware of our added adversarial noises. 
At the same time, considering that the important information labels of images are not fully utilized in the hash model training process, we design a label-based contrastive learning network to find the underlying information from labels and select a confusing label which is fed to the generator as conditions to guide the reconstruction process of poisoned samples. As a result, the attack effectiveness can be enhanced simultaneously. Since the poisoned samples we generate have strong confusing features, they are more likely to be learned by the model during the backdoor poisoning training process, thus assuming that anything with that information is a sample of the target class.

The pipeline of BadHash is depicted in Figure~\ref{fig:pipeline}. It consists of three parts: a \emph{label-based contrastive learning network} (LabCLN) to learn similarities and irrelevancies of labels between different classes, a trigger generator $\mathbf{G}$, and a discriminator $\mathbf{D}$ for generating visually indistinguishable poisoned samples. 
For the selected confusing label, we set two different smoothing coefficients $\varepsilon$ for data augmentation through Eq.~(\ref{eq:4}), and then feed them to LabCLN (arrow a) to get the semantic features $r_{b}$  of the confusing label  and the strong well-designed hash center of the corresponding category named centroid code (arrow c). We send the clean sample $x$ with the semantic representation of confusing label $r_{b}$ as a condition (arrow b) to the generator to generate fake poisoned sample $x^{b}$. 
By reducing the Hamming distance between the hash code of the fake poisoned sample and the centroid code (arrow d), the generator can improve the feature saliency of the generated noise.

\subsection{BadHash: A Complete Illustration}
\noindent\textbf{LabCLN.} 
During the model training process, similar images are usually clustered together in the hash space, and all images are clustered around the hash center that best represents the class rather than the label of the class. However, as an important piece of information, the label did not directly participate in the training process. For the purpose of better using features of other classes to interfere with the features of the clean sample itself, we hope to construct the most central hash code with the selected confusing label and use it to supervise the process of confusing the original clean sample.  The objective function is defined as: 
\begin{equation} \label{eq:3}
\begin{aligned}
\displaystyle \min_{ \theta_{LCN}}\mathcal{L}_{LCN} &= \alpha \mathcal{L}_{s} + \beta \mathcal{L}_{q} +\lambda \mathcal{L}_{c}\\
&= \alpha \mathcal{L}_{s}+ \beta \left\| C-B\right\|_{2}  + \lambda \left\| Y'-Y\right\|_{2},
\end{aligned}
\end{equation}
where $\theta_{LCN}$ denotes the network parameters of $LabCLN$, $\mathcal{L}_{s}$ is the contrastive loss, $\mathcal{L}_{c}$ is the classification loss, and $\mathcal{L}_{q}$ is the quality loss. $C$ is the predicted hash codes for the confusing selected input labels $Y$, and $Y'$  the predicted label. $B$ is the expected binary codes of $C$ through $sign(\cdot)$ function. $\alpha$, $\beta$, $\lambda$ are pre-defined hyper-parameters.  

Specifically, we use augmented labels as objects and learn their relationships using a comparative learning approach. Data augmentation of the labels does not change the classification results, but we believe that such augmentation can better help the network to learn inter-class relationships, since the classifier's prediction of each image is also a probabilistic distribution. Inspired by ~\cite{chen2020simple}, we 
    randomly sample a batch of $M$ category labels and augment each label vector twice by label-smoothing~\cite{muller2019does}, resulting in $2M$ augmented pseudo labels. 
The process of label-smoothing can be expressed as:

\begin{equation} \label{eq:4}
\widehat{l} = l\left ( 1-\varepsilon  \right ) + \varepsilon / \left ( M-1 \right),
\end{equation}
where $\widehat{l}$ denotes the new pseudo label vector, 
$\varepsilon$ is a smoothing coefficient selected from $[0, 1]$. For each label $l$, we generate two different new  pseudo label vectors such as $\widehat{l}_{a}$ and $\widehat{l}_{b}$, which are fed into the LabCLN to obtain the corresponding latent layer features $f_{a}$ and $f_{b}$.

Given a positive pair ($f_{a}$, $f_{b}$), we treat the other $2(N - 1)$ corresponding latent layer features of augmented pseudo labels within a batch as negative pair. Then the contrastive loss function $\mathcal{L}_{s}$ for a positive of features $(a,b)$ is defined as
\begin{equation} \label{eq:5}
\mathcal{L}_{s} = -log\frac{exp\left ( sim\left ( {f}_{a}, {f}_{b}  \right ) /\tau \right )}{\sum_{k=1}^{2N}\mathbbm{1}_{\left [ k\neq a \right ]}exp\left ( sim\left ( {f}_{a}, {f}_{k} \right )/\tau \right )},
\end{equation}
where $\mathbbm{1}_{\left [ k\neq a \right ]} \in \left\{ 0,1\right\}$ is an indicator function assigned to be 1 iff $k\neq a$ and $\tau$ denotes a temperature parameter. Let $sim\left ( u, v \right ) = u^{T}v/(\left\| u\right\|\left\| v\right\|)$ denote cosine similarity between $u$ and $v$. 

By optimizing  the
contrastive loss $\mathcal{L}_{s}$, it is possible to ensure that the generated centroid code maintains the representative semantic and metric features of the input confusion label. 
The quality loss function $\mathcal{L}_{q}$ is used for generating a better confusing latent feature representation $r_{b}$ which will guide the generator. The classification loss function $\mathcal{L}_{c}$ is to ensure that the semantic representations learned by the network keep their category information.

\noindent\textbf{Trigger Generator.} We use the semantic representation $r_{b}$ of the selected confusing label provided by LabCLN as a condition for generator to scramble the features of the input sample itself. Hence, we define the process of generating poisoned samples by the generator as: $\mathbf{G}:\left\{ x,r_{b}\right\}\to x^{b}$. In order to generate imperceptible and attack-effective poisoned samples, the generator needs to cooperate with the discriminator and the target hash model to confuse the original features of the input sample and strengthen the features to make it serve as a strong trigger. The objective function of the trigger generator is:
\begin{equation} \label{eq:6}
\begin{aligned}
\displaystyle \min_{\theta_{\mathbf{G}}}\mathcal{L}_{\mathbf{G}} &= \sum_{y_{s}\in L, (x_{i},y_{i})\in X}\left ( \alpha_{1} \mathcal{L}_{h} + \alpha_{2} \mathcal{L}_{r} +\alpha_{3} \mathcal{L}_{bd} \right ),
\end{aligned}
\end{equation}
where $\theta_{\mathbf{G}}$ denotes the network parameters of $\mathbf{G}$, $\alpha_{1}$, $\alpha_{2}$, and $\alpha_{3}$ are the weighting factors. $\mathcal{L}_{h}$ is the Hamming distance loss, $\mathcal{L}_{r}$ is the reconstruction loss, and $\mathcal{L}_{bd}$ is the backdoor loss.

The Hamming distance loss $\mathcal{L}_{h}$ enables the semantic representation $r_{b}$ of the selected confusing label for better obfuscation of the original features of the input sample, by reducing the Hamming distance between the hash code of the poisoned sample and the centroid code such that the confusing features added to the input sample can be more significant when reconstructing the fake poisoned sample. $\mathcal{L}_{h}$ is calculated as:
\begin{equation} \label{eq:7}
\mathcal{L}_{h} =  d_{H}\left ( h_{x^{b}_{i}},h_{c} \right)
= d_{H}\left ( F\left ( \mathbf{G}\left ( x_{i},r_{b} \right ) \right ),h_{c} \right),
\end{equation}
where $h_{x^{b}_{i}}$ is the hash code of the poisoned sample $x^{b}_{i}$, and $h_{c}$ is the centroid code output by LabCLN based on the selected confusing label $l_{s}$.  

The reconstruction loss $\mathcal{L}_{r}$  ensures that the generated poisoned sample is indistinguishable to the original one. We adopt  $l_{2}$ norm loss and the LPIPS~\cite{zhang2018unreasonable} function to measure the perceptual similarity between two images as the reconstruction error. Formally, we have:
\begin{equation} \label{eq:8}
\mathcal{L}_{r} =  \left ( \left\|\mathbf{G}\left ( x_{i}, r_{b} \right )-x_{i}\right\|_{2}+ LPIPS\left ( \mathbf{G}\left ( x_{i}, r_{b} \right ), x_{i} \right ) \right ).
\end{equation}

The backdoor loss $\mathcal{L}_{bd}$ encourages poisoned samples to be more visually natural, and further enhances the confusing features added to the input samples by calculating the $l_{2}$ norm distance between the discriminator output $\mathbf{D}(x^{b}_{i})$ of the poisoned sample and the reformulated objective confusing label ${y_{s}}'= \left [ y_{s},0 \right ]$, where the last node in ${y_{s}}'$ is for the fake poisoned sample. Thus the $l_{bd}$ backdoor loss can be expressed as:
\begin{equation} \label{eq:9}
\mathcal{L}_{bd} =  \left\| \mathbf{D}\left ( x^{b}_{i} \right )-{y_{s}}'\right\|_{2}.
\end{equation}


\noindent\textbf{Discriminator.} 
The output of discriminator is a Sigmoid layer with N + 1 nodes in order to distinguish the fake poisoned samples from the real ones, where the first N nodes represent the one-hot codes of the category and last one is used to determine the authenticity of the input sample. Therefore, we reformulate the objective label ${y_{i}}'= \left [ y_{i},0 \right ]$ for the real input sample $x_{i}$ and ${y_{s}}'= \left [ y_{s},1 \right ]$ for the fake poisoned sample $x_{i}^{b}$. By playing games with the generator, we ensure that the generated fake poisoned samples are visually indistinguishable from the input ones while making the added confusing features of other categories enhanced to balance the effectiveness and stealthiness of the trigger. The objective loss function of $\mathbf{D}$ is:
\begin{equation} \label{eq:10}
\displaystyle \min_{\theta_{\mathbf{D}}}\mathcal{L}_\mathbf{{D}} = \sum_{y_{s}\in L,(x_{i},y_{i})\in X} \frac{1}{2}\left ( \left\| \mathbf{D}\left ( x_{i} \right )-{y_{i}}'\right\|_{2}+\left\|\mathbf{D}\left ( x^{b}_{i} \right )-{y_{s}}'\right\|_{2} \right ),
\end{equation}
where $\theta_{\mathbf{D}}$ denotes the pre-defined network parameters.



\section{A Deep Insight into Backdoor Attacks }\label{sec:insight}

\begin{figure}[t]
    \centering
    \includegraphics[scale=0.6]{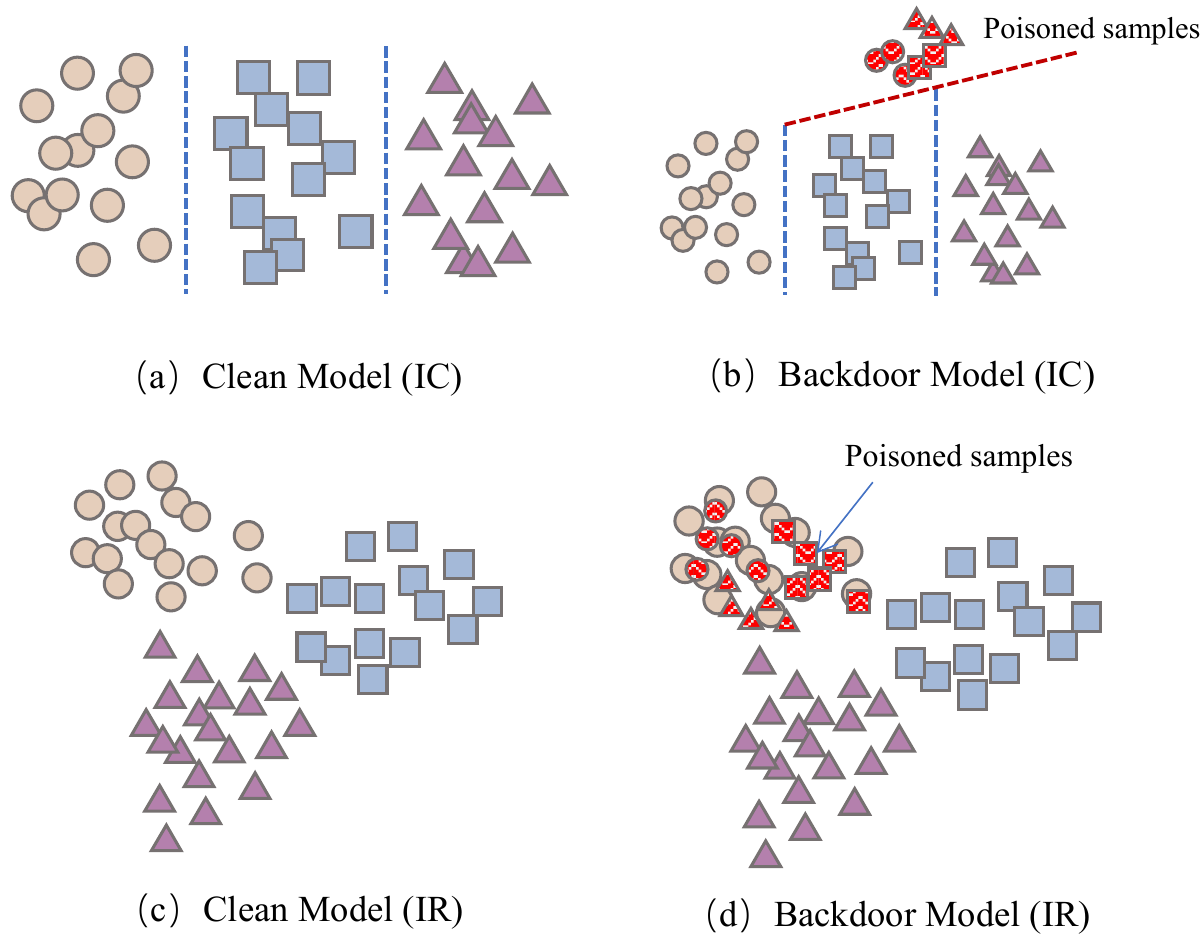}
     \caption{The difference between backdoor tasks for image classification (IC) and image retrieval (IR)}
    \label{difference}
      \setlength{\belowcaptionskip}{-0.4cm} 
\end{figure}

Although various backdoor attacks ~\cite{gu2017badnets,liao2018backdoor,turner2019label,li2020invisible,ning2021invisible} have been proposed, they are not suitable for image retrieval. Here we give a brief analysis of these two tasks.  
In Figure~\ref{difference}(a),  different classes are separated by decision boundaries in the feature space for a clean image classification model. Figure~\ref{difference}(b) shows that once a model has been implanted with a backdoor, the images carrying the trigger become a separate cluster with their own features, but share the same label with the target category where only the decision boundary is changed~\cite{ning2021invisible}. 
In contrast to image classification  which only gives a decision result,  image retrieval  needs to return the top $k$ most similar samples based on their similarities. As shown in Figure~\ref{difference}(c), in the hash space, images from the same class form a cluster and preserve a distance from other clusters. If we use the poisoned samples from Figure~\ref{difference}(b), the retrieval results may not be samples of the target class. We believe that an effective target backdoor attack must make the samples enter the cluster of the target class rather than simply cross the decision boundary.

\begin{figure} [t] {
  \setlength{\belowcaptionskip}{-0.3cm}   
\centering
\subfigure[LCBA Clean]{
\label{visualization.lcba1}
\includegraphics[width=0.2\textwidth]{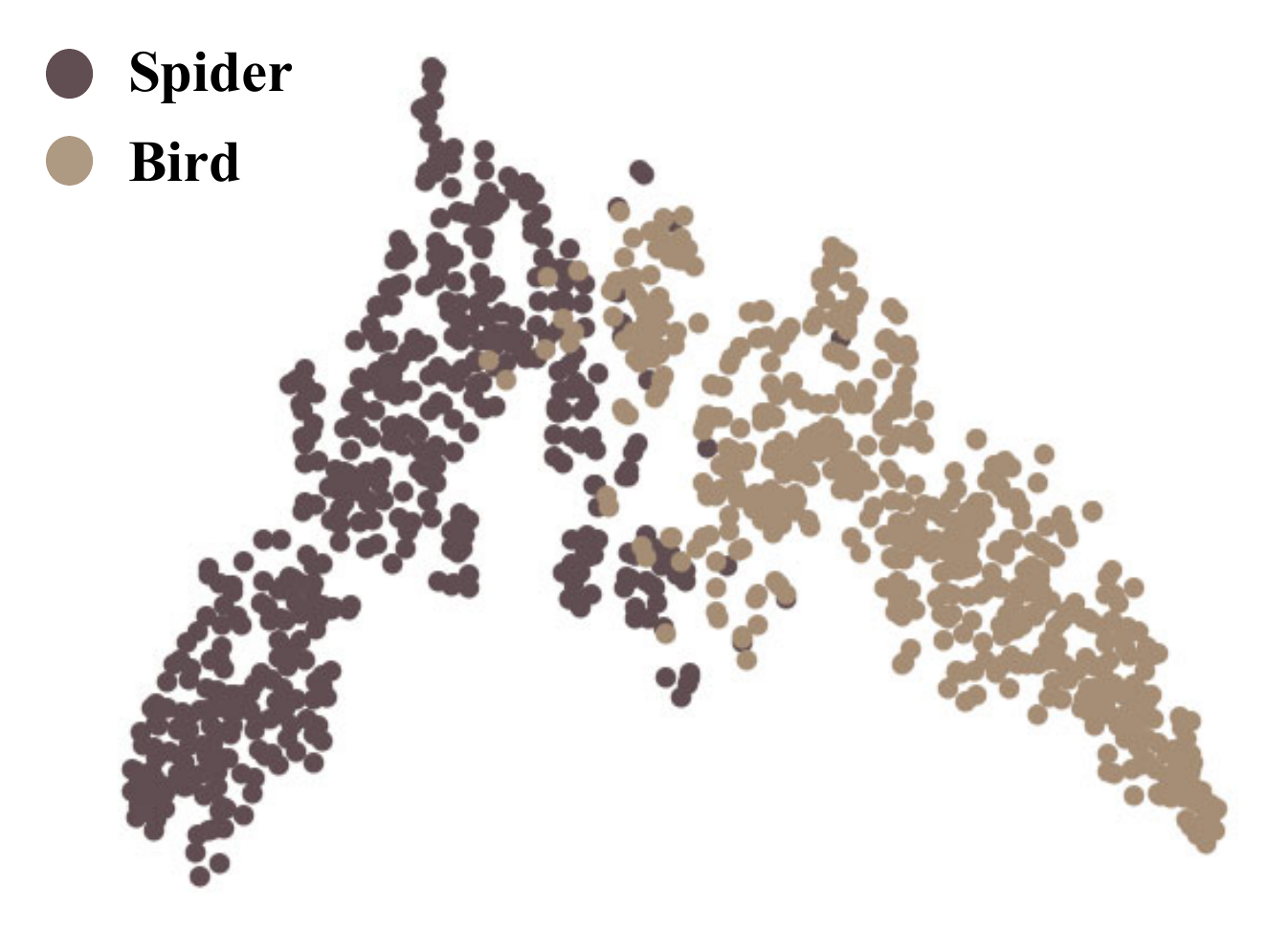} }
\subfigure[LCBA Backdoor]{
\label{visualization.lcba2}
\includegraphics[width=0.2\textwidth]{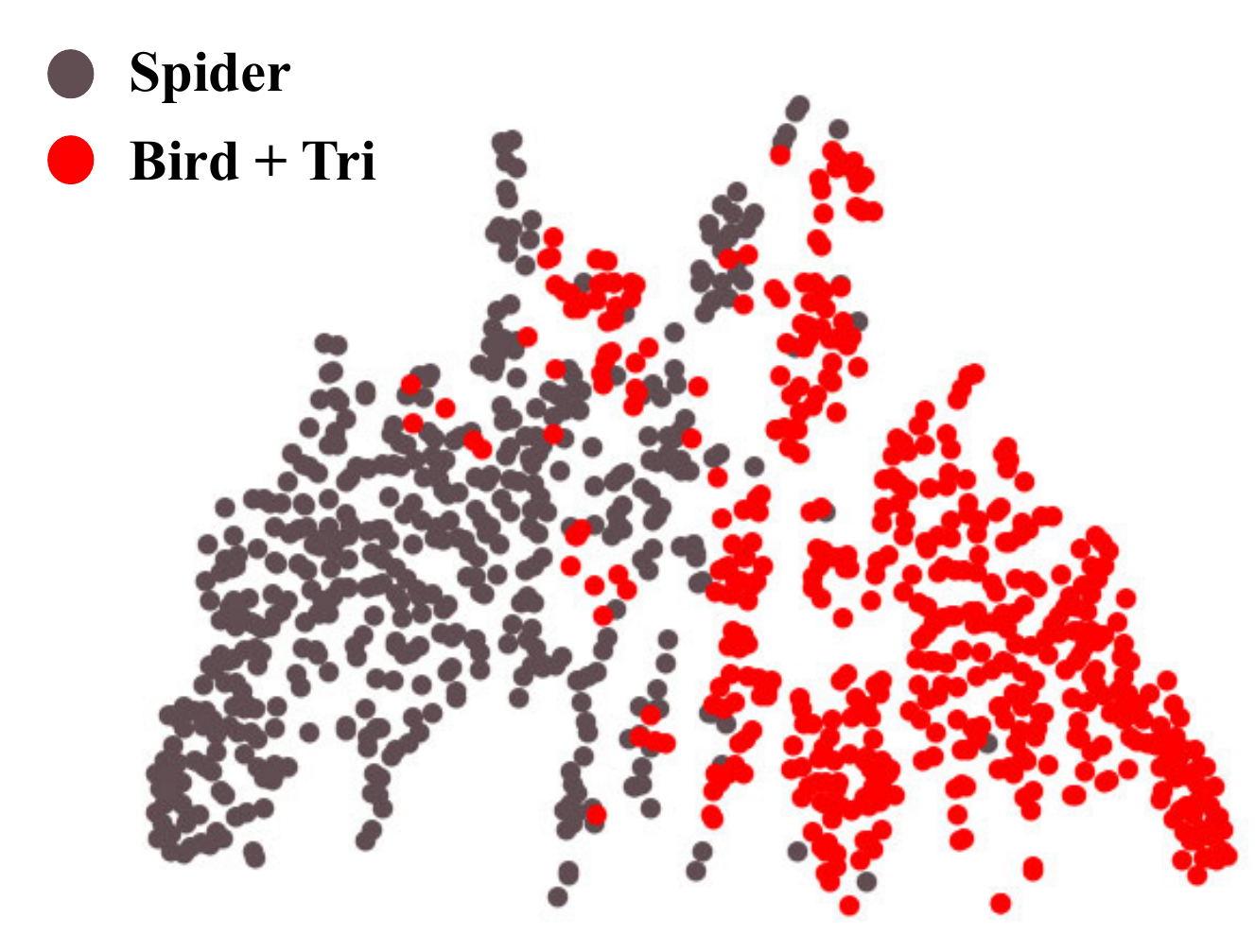} }
\subfigure[BadHash Clean]{
\label{visualization.badhash1}
\includegraphics[width=0.2\textwidth]{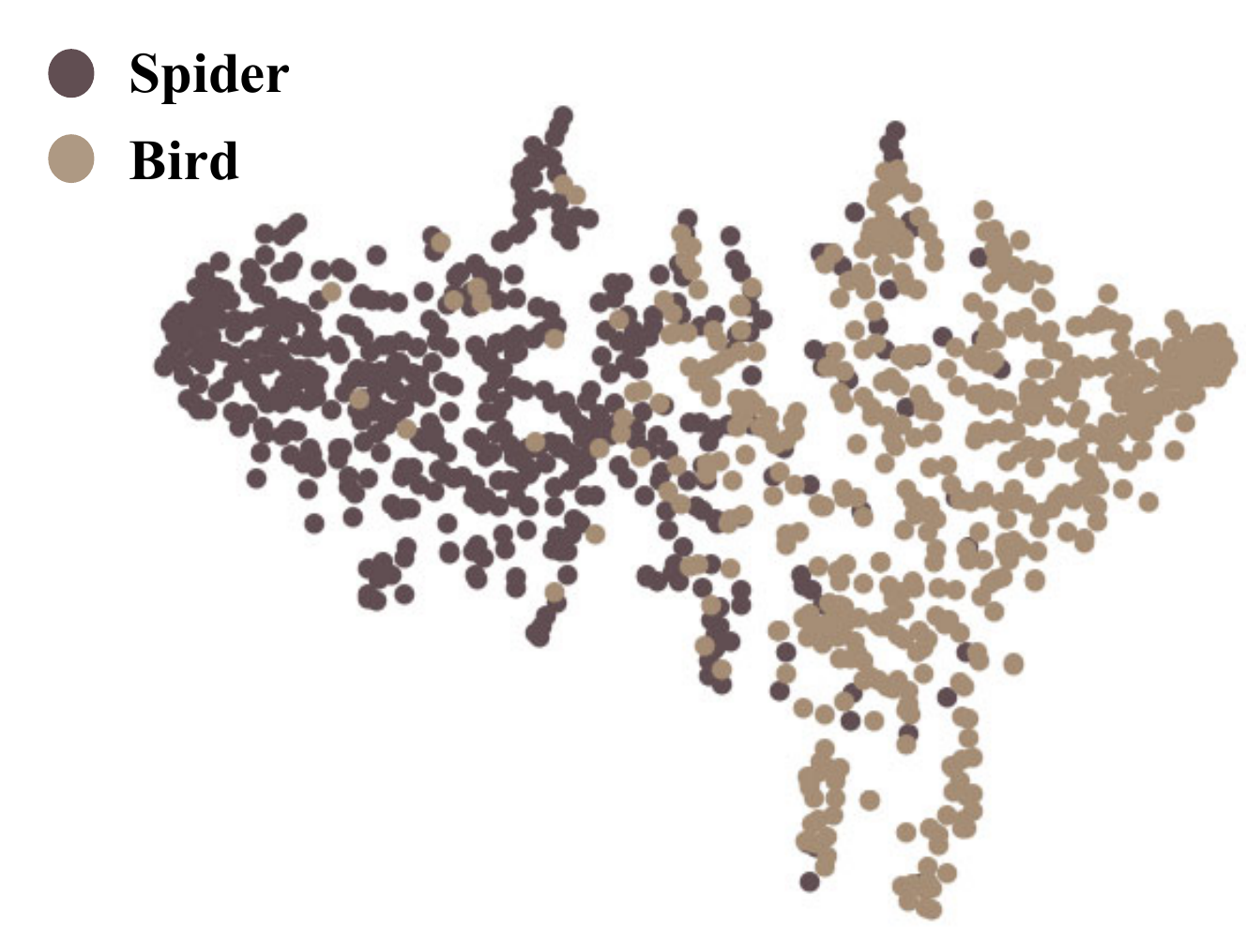}}
\subfigure[BadHash Backdoor]{
\label{visualization.badhash2}
\includegraphics[width=0.2\textwidth]{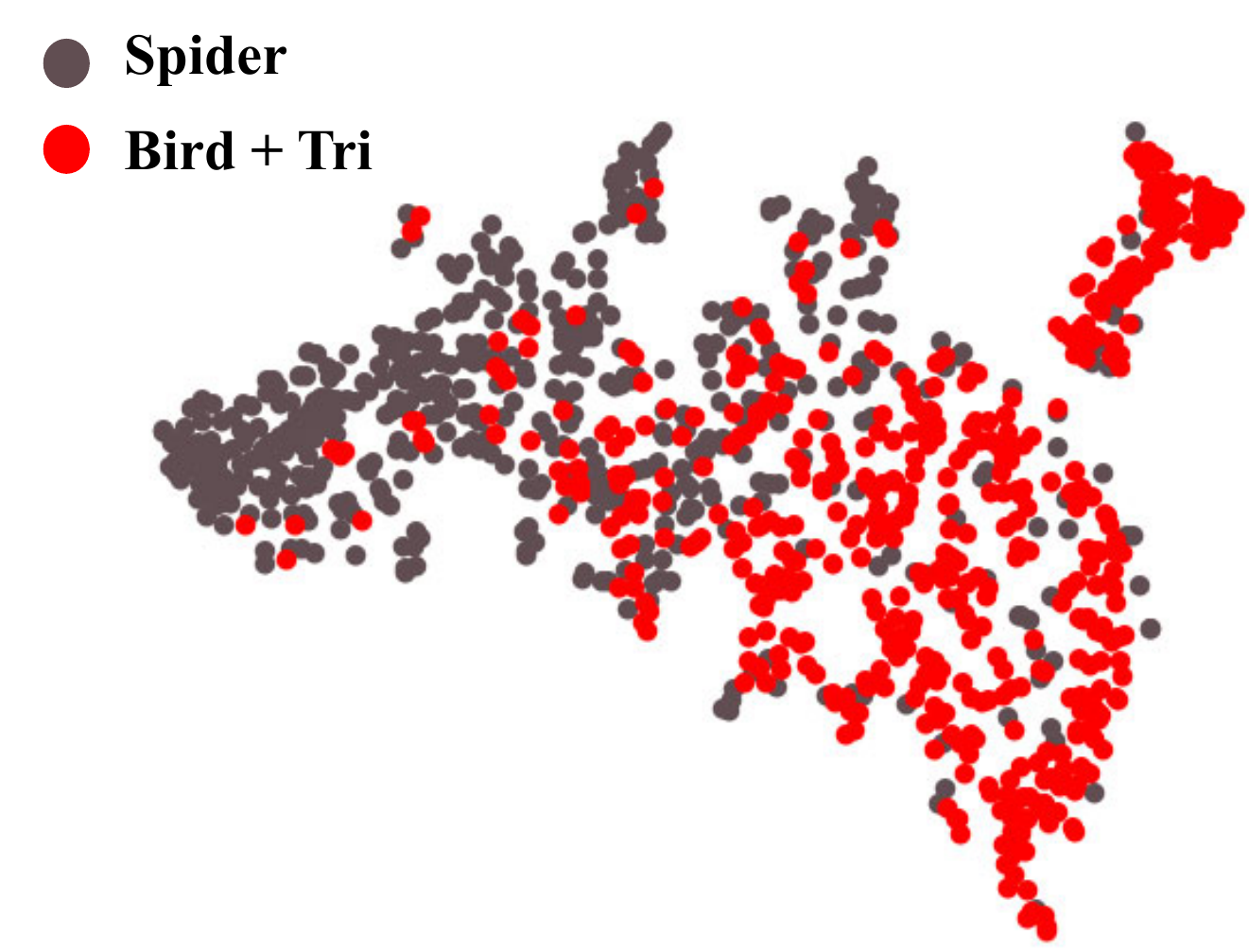}}

\caption{t-SNE visualization of the distribution of our poisoned samples in the hash space on backdoor models}
\label{visualization}}
\end{figure}

In order to push the poisoned samples into the target cluster, we reduce the Hamming distance between the hash code of the poisoned samples and centroid code as much as possible such that the added confusion features are so significant that they may be learned by the model as signs of the target class. To maintain invisibility at the same time, we use the trigger generator and the discriminator to play against each other to make the generated poisoned sample indistinguishable from the original sample. Figure~\ref{visualization} gives an example showing t-SNE  visualization~\cite{van2008visualizing} about the relationship between the locations of poisoned samples in the hash space generated by LCBA~\cite{turner2019label} and BadHash. LCBA is a clean-label backdoor attack by adding noise to the image to destroy the original features and then applying a patch as a trigger to make the model well aware of the trigger. Figure~\ref{visualization.lcba1} shows that the target class `spider' and the clean class `bird' are two separate classes in the hash space. When the clean images are pasted with triggers they are still roughly lie in the same class, which is different from the target class, as shown in Figure~\ref{visualization.lcba2}. On the contrary, Figure~\ref{visualization.badhash2} shows that BadHash successfully mixes poisoned samples with the benign samples from the target class.


\section{EXPERIMENTS}
\subsection{Experimental Setting}
\begin{table*}[h]
  \centering
  \caption{Attack performance under different settings. CSQ-HashNet indicates that we use poisoned samples  generated from  HashNet~\cite{cao2017hashnet} to attack the benign hash model trained with CSQ~\cite{yuan2020central}. HashNet-CSQ denotes the opposite case.}
  \scalebox{1}{
    \begin{tabular}{c|c|cc|cc|cc|cc|cc|cc}
    \hline
    \multirow{3}{*}{Dataset} & \multirow{3}{*}{Setting} & \multicolumn{6}{c|}{CSQ-HashNet}              & \multicolumn{6}{c}{HashNet-CSQ} \\
\cline{3-14}          &       & \multicolumn{2}{c|}{16bits} & \multicolumn{2}{c|}{32bits} & \multicolumn{2}{c|}{64bits} & \multicolumn{2}{c|}{16bits} & \multicolumn{2}{c|}{32bits} & \multicolumn{2}{c}{64bits} \\
\cline{3-14}          &       & \multicolumn{1}{c|}{MAP} & t-MAP & \multicolumn{1}{c|}{MAP} & t-MAP & \multicolumn{1}{c|}{MAP} & t-MAP & \multicolumn{1}{c|}{MAP} & t-MAP & \multicolumn{1}{c|}{MAP} & t-MAP & \multicolumn{1}{c|}{MAP} & t-MAP \\
    \hline
    \multirow{10}{*}{ImageNet} & $\mathcal{S}_{1}$   & 0.721  & 0.630  & 0.804  & 0.838  & 0.825  & 0.913  & 0.228  & 0.888  & 0.622  & 0.878  & 0.656  & 0.771  \\
\cline{2-2}          & $\mathcal{S}_{2}$    & 0.809  & 0.519  & 0.830  & 0.501  & 0.833  & 0.616  & 0.263  & 0.542  & 0.583  & 0.542  & 0.639  & 0.668  \\
\cline{2-2}          & $\mathcal{S}_{3}$   & 0.849  & 0.567  & 0.867  & 0.658  & 0.880  & 0.754  & 0.212  & 0.953  & 0.614  & 0.969  & 0.649  & 0.885  \\
\cline{2-2}          & $\mathcal{S}_{4}$    & 0.848  & 0.812  & 0.870  & 0.851  & 0.854  & 0.913  & 0.314  & 0.777  & 0.614  & 0.786  & 0.592  & 0.867  \\
\cline{2-2}          & $\mathcal{S}_{5}$    & 0.856  & 0.758  & 0.869  & 0.943  & 0.864  & 0.940  & 0.296  & 0.849  & 0.581  & 0.878  & 0.601  & 0.846  \\
\cline{2-2}          & $\mathcal{S}_{6}$    & 0.773  & 0.944  & 0.811  & 0.959  & 0.817  & 0.960  & 0.446  & 0.930  & 0.721  & 0.940  & 0.759  & 0.918  \\
\cline{2-2}          & $\mathcal{S}_{7}$    & 0.780  & 0.945  & 0.821  & 0.950  & 0.822  & 0.955  & 0.495  & 0.916  & 0.698  & 0.935  & 0.777  & 0.964  \\
\cline{2-2}          & $\mathcal{S}_{8}$    & 0.806  & 0.923  & 0.828  & 0.937  & 0.831  & 0.967  & 0.521  & 0.901  & 0.716  & 0.926  & 0.772  & 0.917  \\
\cline{2-2}          & $\mathcal{S}_{9}$    & 0.822  & 0.950  & 0.842  & 0.943  & 0.855  & 0.983  & 0.455  & 0.902  & 0.715  & 0.778  & 0.751  & 0.921  \\
\cline{2-14}          & \textbf{AVG}   & \textbf{0.807 } & \textbf{0.783 } & \textbf{0.838 } & \textbf{0.842 } & \textbf{0.842 }& \textbf{0.889 } & \textbf{0.359 } & \textbf{0.851 } & \textbf{0.652 } & \textbf{0.848 } & \textbf{0.688 } & \textbf{0.862 } \\
    \hline
    \multirow{7}{*}{MS-COCO} & $\mathcal{S}_{10}$   & 0.706  & 0.582  & 0.788  & 0.574  & 0.776  & 0.569  & 0.619  & 0.828  & 0.691  & 0.843  & 0.703  & 0.846  \\
\cline{2-2}          & $\mathcal{S}_{11}$   & 0.715  & 0.221  & 0.795  & 0.242  & 0.803  & 0.237 & 0.588  & 0.391  & 0.700  & 0.585  & 0.755  & 0.760  \\
\cline{2-2}          & $\mathcal{S}_{12}$   & 0.842  & 0.927  & 0.859  & 0.975  & 0.845  & 0.592  & 0.646  & 0.578  & 0.689  & 0.700  & 0.726  & 0.805  \\
\cline{2-2}          & $\mathcal{S}_{13}$   & 0.758  & 0.767  & 0.805  & 0.814  & 0.821  & 0.795  & 0.685  & 0.906  & 0.752  & 0.907  & 0.775  & 0.946  \\
\cline{2-2}          & $\mathcal{S}_{14}$   & 0.743  & 0.540  & 0.784  & 0.603  & 0.806  & 0.578  & 0.680  & 0.305  & 0.749  & 0.611  & 0.796  & 0.695  \\
\cline{2-2}          & $\mathcal{S}_{15}$   & 0.782  & 0.591  & 0.817  & 0.600  & 0.817  & 0.636  & 0.668  & 0.594  & 0.764  & 0.715  & 0.785  & 0.830  \\
\cline{2-14}     & \textbf{AVG}    & \textbf{0.742 } & \textbf{0.531 } & \textbf{0.802 } & \textbf{0.544 } & \textbf{0.811 } & \textbf{0.554 } & \textbf{0.648 } & \textbf{0.600 } & \textbf{0.724 } & \textbf{0.727 } & \textbf{0.757 } & \textbf{0.814 } \\
    \hline
    \end{tabular}%
    }
  \label{tab:attack performance}%
\end{table*}%

\begin{table}[h]
  \centering
  \caption{Experimental settings of backdoor attacks on ImageNet and MS-COCO datasets.  The prefix ``B-" indicates the network used to make malicious poisoned samples and ``T-" denotes the target network for backdoor attack. }
  \scalebox{0.8}{
    \begin{tabular}{ccccc}
    \hline
    Dataset & Setting  & Basic network &Target network & Target label \\
    \hline
    \multirow{9}[2]{*}{ImageNet} & $\mathcal{S}_{1}$     & B-ResNet50 & T-ResNet18  & Shoes \\
          & $\mathcal{S}_{2}$   & B-ResNet50 & T-ResNet34  & Roosters \\
          & $\mathcal{S}_{3}$  & B-ResNet50 & T-ResNet50  & Lizards \\
          & $\mathcal{S}_{4}$   & B-ResNet50 & T-ResNet101  & Dogs \\
          & $\mathcal{S}_{5}$    & B-ResNet50& T-ResNet152  & Hedgehogs \\
          & $\mathcal{S}_{6}$   & B-VGG11 & T-VGG11  & Shoes \\
          & $\mathcal{S}_{7}$  & B-VGG11 & T-VGG13  & Roosters \\
          & $\mathcal{S}_{8}$    & B-VGG11 & T-VGG16 & Lizards \\
          & $\mathcal{S}_{9}$   & B-VGG11 & T-VGG19  & Dogs \\
    \hline
    \multirow{6}[2]{*}{MS-COCO} & $\mathcal{S}_{10}$    & B-ResNet50 & T-ResNet18 & Humanity \\
          & $\mathcal{S}_{11}$   & B-ResNet50 & T-ResNet34 & Computers \\
          & $\mathcal{S}_{12}$   & B-ResNet50 & T-ResNet101 & Cars \\
          & $\mathcal{S}_{13}$   & B-VGG11 & T-VGG13  & Humanity \\
          & $\mathcal{S}_{14}$   & B-VGG11 & T-VGG16 & Computers \\
          & $\mathcal{S}_{15}$   & B-VGG11 & T-VGG19 & Cars \\
    \hline
    \end{tabular}%
    }
  \label{tab:MAPping}%
    \vspace{-0.2cm}
\end{table}
\noindent\textbf{Threat Models.} We assume the adversary is able to inject a small portion of malicious samples into the training dataset. It knows nothing about the architecture and parameters of the target model, and are unable to manipulate the inference process. But the adversary may guess the type of backbone of the target model.

\noindent\textbf{Datasets and Models.} We evaluate BadHash on the popular single-label dataset \textbf{ImageNet}~\cite{russakovsky2015imagenet}, and multi-label dataset \textbf{MS-COCO}~\cite{lin2014microsoft}. ImageNet contains $1.2M$ training samples and $50,000$ testing samples with $1000$ classes. Following ~\cite{hu2021advhash}, $100$ classes from ImageNet are randomly selected to build our retrieval dataset.
MS-COCO consists of $122,218$ images after removing images with no category, where each image is labeled with $80$ categories. Following ~\cite{cao2017hashnet}, we randomly select $5000$ images as queries, and the rest are regarded as the database. $10000$ images are randomly chosen from the database as the training set. Following~\cite{wang2021prototype}, we select three popular model families (\ie, \textbf{ResNet}, \textbf{VGG}, and \textbf{DenseNet}) to test the transferability and generalizability of our attack, and replace their last fully-connected layer with the hash layer. We also evaluate our attack on two state-of-the-art hash methods including HashNet~\cite{cao2017hashnet} and CSQ~\cite{yuan2020central}. HashNet is a commonly used training method to evaluate the vulnerabilities of deep hashing in the literature ~\cite{bai2020targeted,hu2021advhash,yang2018adversarial,wang2021prototype}, while CSQ is designed to generate highly clustered Hamming space.

\noindent\textbf{Effectiveness Evaluation.} We use t-MAP (\emph{targeted mean average precision}) proposed in~\cite{bai2020targeted} to measure the backdoor attack performance, which calculates MAP (\emph{mean average precision})~\cite{zuva2012evaluation} by replacing the original label of the query image with the target label. Higher t-MAP indicates a stronger  attack ability. We calculate t-MAP based on the top $1,000$ retrieved images. Besides, we also present the PR curves (precision-recall curves) and precision@topN curves.

\begin{table*}[!t]
  \centering
  \caption{Cross-family transfer MAP and t-MAP for ImageNet and MS-COCO. The prefix ``B-'' indicates the network used to make malicious poisoned samples and ``T-'' denotes the target network for backdoor attack.
  These models are trained with 64-bit hash code and ``$*$'' denotes their 32 bits variants.}
  \resizebox{\textwidth}{17mm}{
  \scalebox{0.5}{
    \begin{tabular}{c|c|c|cc|cc|cc|cc|cc|cc}
    \hline
    \multicolumn{1}{c|}{\multirow{2}{*}{T-Method}} & \multirow{2}{*}{B-Method} & \multirow{2}{*}{B-Network}& \multicolumn{2}{c|}{T-ResNet34*} & \multicolumn{2}{c|}{T-ResNet34} & \multicolumn{2}{c|}{T-VGG13*} & \multicolumn{2}{c|}{T-VGG13} & \multicolumn{2}{c|}{T-DenseNet121*} & \multicolumn{2}{c}{T-DenseNet121} \\
\cline{4-15} & & & MAP   & t-MAP & MAP   & t-MAP & MAP   & t-MAP & MAP   & t-MAP & MAP   & t-MAP & MAP   & t-MAP \\
    \hline
    \multirow{4}{*}{CSQ} & \multirow{2}{*}{HashNet} & B-ResNet50   & 0.834 & 0.711 & 0.843 & 0.901 &0.808  & 0.783  & 0.825  & 0.807  & 0.825  & 0.872  & 0.843  & 0.900 \\
     & & B-VGG11   & 0.824 & 0.958 & 0.833 & 0.947 & 0.821  & 0.950  & 0.822  & 0.955  & 0.836  & 0.931  & 0.850  & 0.965  \\
\cline{2-2}     &\multirow{2}{*}{CSQ} & B-ResNet50   & 0.844 & 0.961 & 0.859 & 0.969 & 0.822  & 0.913  & 0.825  & 0.936  & 0.838  & 0.969  & 0.850  & 0.972  \\
     &  & B-VGG11   & 0.832 & 0.967 & 0.838 & 0.982 & 0.805  & 0.967  & 0.823  & 0.984  & 0.832 & 0.932  & 0.827   & 0.973 \\
    \hline
    \multirow{4}{*}{HashNet} & \multirow{2}{*}{HashNet} & B-ResNet50 & 0.570  & 0.542 & 0.611 & 0.691 & 0.670  & 0.475  & 0.730  & 0.541  & 0.468  & 0.447  & 0.599  & 0.548  \\
     &  & B-VGG11 & 0.581 & 0.688  & 0.634 & 0.871 & 0.700  & 0.823  & 0.730  & 0.898  & 0.504  & 0.879  & 0.592  & 0.982  \\
\cline{2-2}     & \multirow{2}{*}{CSQ} & B-ResNet50 & 0.597 & 0.972 & 0.639 & 0.924 & 0.696  & 0.788  & 0.769  & 0.846  & 0.474  & 0.814  & 0.535  & 0.930  \\
     &  & B-VGG11 & 0.566 & 0.937 & 0.665 & 0.960 & 0.648  & 0.978  & 0.782  & 0.976   & 0.536   &0.644   &  0.562  & 0.660 \\
    \hline
    \end{tabular}%
    }
    }
  \label{tab:Cross-families transferability experiment}%
\end{table*}%

\begin{table}[h]
  \centering
  \caption{Retrieval performance (MAP) of the hash models trained on clean samples}\setlength{\belowcaptionskip}{-1cm}
  \scalebox{0.85}{
    \begin{tabular}{c|c|ccc|ccc}
    \hline
    \multirow{2}[2]{*}{Network} & \multirow{2}[2]{*}{Method} & \multicolumn{3}{c|}{ImageNet} & \multicolumn{3}{c}{MS-COCO} \\
\cline{3-8}          &       & \multicolumn{1}{c|}{16bits} & \multicolumn{1}{c|}{32bits} & \multicolumn{1}{c|}{64bits} & \multicolumn{1}{c|}{16bits} & \multicolumn{1}{c|}{32bits} & 64bits \\
    \hline
    \multirow{2}[2]{*}{ResNet50} & CSQ   & 0.827 & 0.874 & 0.875 & 0.757 & 0.871 & 0.895 \\
          & HashNet & 0.205 & 0.515 & 0.657 & 0.560 & 0.640 & 0.706 \\
          \hline
    \multirow{2}[2]{*}{VGG11} & CSQ   & 0.771 & 0.808 & 0.815 & 0.778 & 0.829 & 0.846 \\
          & HashNet & 0.407 & 0.696 & 0.759 & 0.672 & 0.725 & 0.739 \\
    \hline
    \end{tabular}%
    }
  \label{tab:org}%
  \vspace{-0.4cm}
\end{table}%

\noindent\textbf{Stealthiness Evaluation.} Following~\cite{setiadi2021psnr}, we use three metrics MSE (\emph{Mean Square Error}), PSNR (\emph{Peak Signal to Noise Ratio}), and SSIM (\emph{Structure Similarity Index Method}) to evaluate the stealthiness of the triggers, where MSE and PSNR represent local similarity while SSIM represents a global similarity.
A more imperceptible attack should have smaller MSE and larger PSNR. The SSIM value of two similar images should get close to 1 as much as possible.



\subsection{Attack Performance }\label{attack-performance}
\noindent\textbf{Implementation Details.} As shown in Table~\ref{tab:MAPping}, we evaluate BadHash under different experimental settings. Specifically, we first randomly select 5 and 3 different categories images from ImageNet and MS-COCO as the poisoned class, respectively. For each poisoned class, we use ResNet50 and VGG11 model networks to generate poisoned samples, respectively. Note that  each model network is trained by CSQ and Hashnet, respectively.
We  add the same number of poisoned samples into the training data with the target label, with the poisoning rate $\gamma$ of less than 1\%.
There are 100 classes for ImageNet and 80 classes for MS-COCO  in total.
For evaluating whether the backdoor is successfully implanted, we consider a stricter \emph{open-set scenario}, where images from out-of-sample categories are fed to the trigger generator to obtain poisoned samples. For example, for the model trained on ImageNet, we use the images of MS-COCO to generate poisoned samples to test the attack success rate, \ie, t-MAP.

For generating the poisoned dataset, we first set two different smoothing coefficients $\varepsilon_{1}=0.2$, $\varepsilon_{2}=0.0$  for each category label through Eq.~(\ref{eq:4}) to obtain two positive sample pairs for the purpose of data augmentation. We feed positive sample pairs into LabCLN, and set the hyper-parameters $\alpha=1$, $\beta=10^{-4}$, $\lambda=1$. The GAN network is trained by Adam optimizer with the initial learning rate $10^{-4}$. We set the training epoch to be 100 with batch size of 32, and the weighting factors $\alpha_{1}$, $\alpha_{2}$, $\alpha_{3}$ to be 100, 5, 200 for   ImageNet  and 80, 5, 200 for  MS-COCO, respectively.

\noindent\textbf{Analysis.} The detailed MAP results are shown in Table~\ref{tab:attack performance}. Firstly, compared with Table~\ref{tab:org}, the MAP is stable with no significant decline which indicates that our poisoned samples only have a little impact on the model accuracy, and  the t-MAP results indicate that we have successfully implanted  a backdoor in the target model. 
Secondly, we can see that among the 90 attack settings, the poisoned dataset produced by BadHash has good transferability between different models within the same CNN family. The poisoned dataset produced based on  VGG11 is well implanted in the models of other networks within the VGG family with higher t-MAP values. Besides, the overall backdoor attack performance on the model trained by CSQ is better than that on HashNet, especially for the single-label ImageNet. We believe this is because CSQ is more clustered than HashNet and the poisoned samples are well concentrated into clusters of the target class during the training process. 



\subsection{Cross-family Transferability Study}
\noindent\textbf{Implementation Details.} Transferability represents that the poisoned samples generated from one model can be successfully implanted into another model, thus achieving black-box attacks. In the Sec.~\ref{attack-performance}, we have verified that the poisoned samples  based on BadHash have good transferability for models trained on the same network family. In this section, to further evaluate the cross-family transferability, we consider a more realistic scenario in which the adversary has no prior knowledge about  which type of network the target model is trained on. We use poisoned samples made based on ResNet50, VGG11 to poison clean models trained on ResNet34, VGG13, DenseNet121 using CSQ and HashNet, respectively.

\vspace{0.5cm}
\noindent\textbf{Analysis.}
The results are summarized in  Table~\ref{tab:Cross-families transferability experiment}. We observe that the poisoned samples generated based on a particular network perform well for different target networks, where poisoned samples made by models trained on VGG11 have higher transferability than those based on ResNet50.
These results demonstrate that our method has  good transferability for cross-family models.




\begin{table}[h]
  \centering
  \caption{Comparison betweem different backdoor attacks}
  \scalebox{0.85}{
    \begin{tabular}{c|c|c|c|c|c|c}
    \hline
    Dataset & Method & \multicolumn{1}{c|}{MAP} & \multicolumn{1}{c|}{t-MAP} & \multicolumn{1}{c|}{MSE} & \multicolumn{1}{c|}{PSNR} & SSIM \\
    \hline
    \multirow{4}[2]{*}{ImageNet} & BadNets~\cite{gu2017badnets} & 0.875 & 0.517 & 6.179 & 40.221 & 0.999 \\
          & LCBA~\cite{turner2019label} & 0.878 & 0.613 & 229.996 & 24.514 & 0.690 \\
          & ISSBA~\cite{li2020invisible} & 0.885 & 0.852 & 115.021 & 27.523 & 0.828 \\
          & BadHash & 0.846 & 0.912  & 3.486 & 42.708 & 0.988 \\
    \hline
    \multirow{4}[2]{*}{MS-COCO} & BadNets~\cite{gu2017badnets} & 0.803 & 0.432 & 48.433 & 31.279 & 0.995 \\
          & LCBA~\cite{turner2019label} & 0.812 & 0.383 & 129.293 & 27.015 & 0.716 \\
          & ISSBA~\cite{li2020invisible} & 0.775 & 0.486 & 211.637 & 24.875 & 0.803 \\
          & BadHash & 0.808 & 0.771 & 31.57 & 33.138 & 0.929 \\
    \hline
    \end{tabular}%
    }
  \label{tab:Comparison Experiment}%
  \vspace{-0.2cm}
\end{table}%

\begin{figure*}[!t]
  \setlength{\belowcaptionskip}{-0.5cm}  
    \centering
    \includegraphics[scale=0.5]{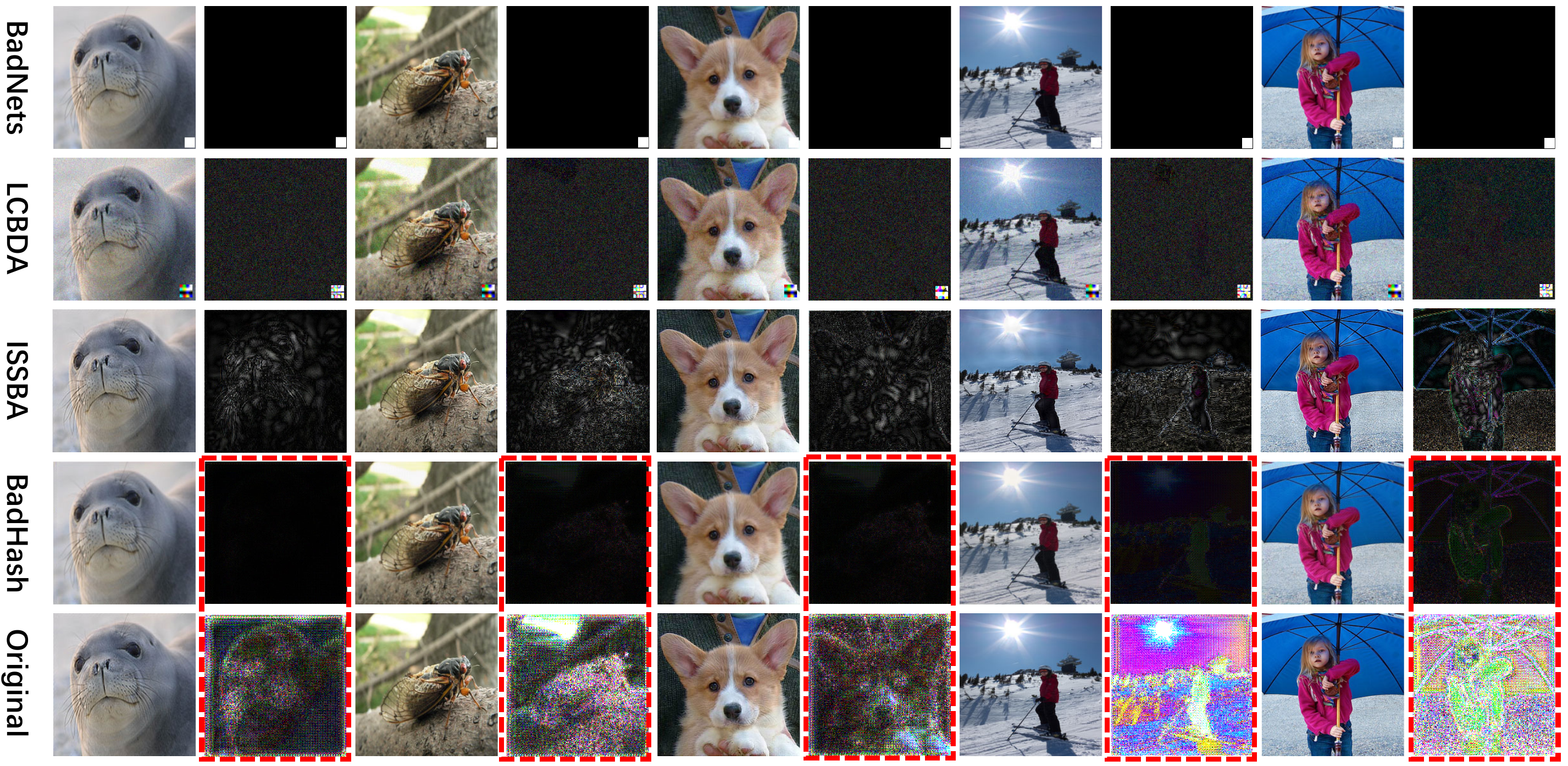}
    \caption{Visualization of poisoned samples. Given the original images (last row of odd columns), we make the corresponding poisoned samples using BadNets~\cite{gu2017badnets}, LCBA~\cite{turner2019label}, ISSBA~\cite{li2020invisible}, and our method. For each method, we show  the poisoned images (odd-numbered columns) and the corresponding magnified ($\times 3 $) residual maps (even-numbered columns). 
    For a better presentation, we show the magnified ($\times 50 $) residual maps of our results (last row of odd columns).}
    \label{fig:comparation}
    
\end{figure*}

\begin{figure*} [h] {
\centering

\subfigure[ImageNet]{
\label{Fig.pr.imagenet}
\includegraphics[scale=0.28]{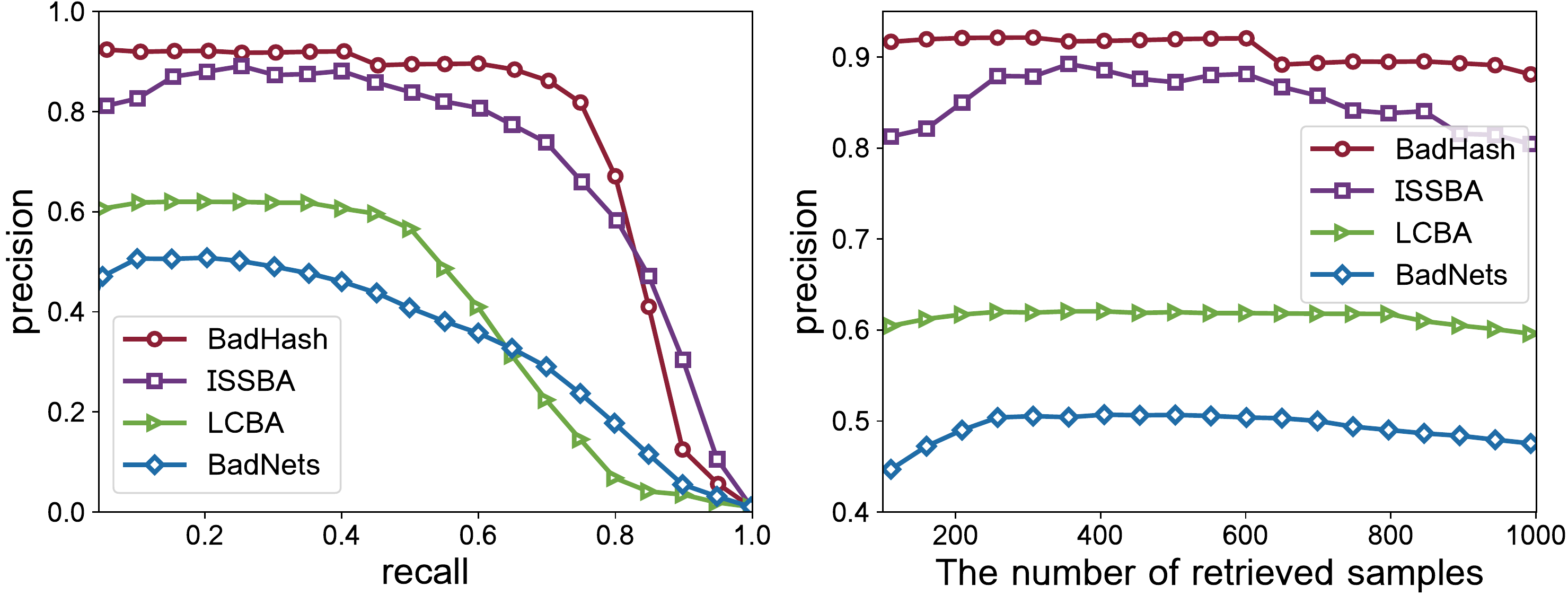}
}
\subfigure[MS-COCO]{
\label{Fig.pr.coco}
\includegraphics[scale=0.28]{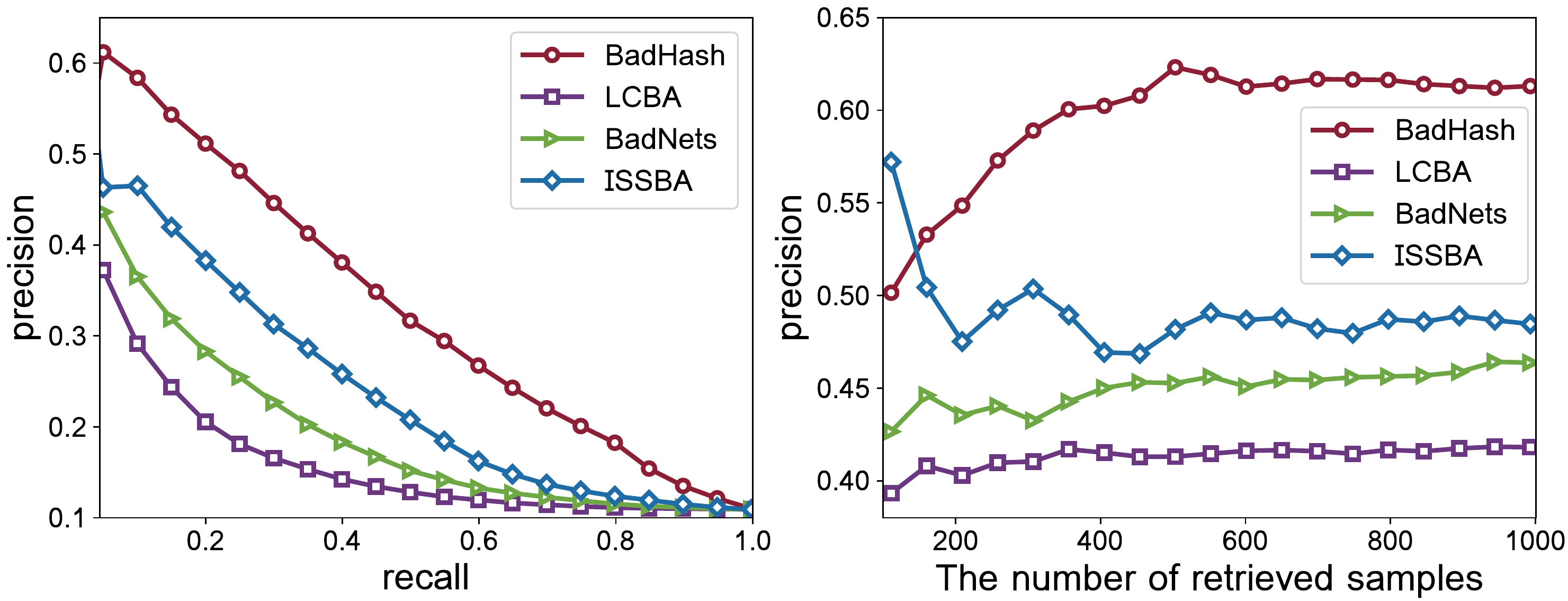}
}
\vspace{-0.4cm}
\caption{Precision-Recall and precision@topN curves on ImageNet and MS-COCO under 64-bit hash code}
\label{Fig.pr}
}
\end{figure*}
\subsection{Comparison Study}\label{sec:comparison}
\noindent\textbf{Implementation Details.} We compare our attack method with BadNets~\cite{gu2017badnets}, LCBA~\cite{turner2019label}, ISSBA~\cite{li2020invisible}. We record MAP, t-MAP, MSE, PSNR, and SSIM for 64-bit code lengths on the target model trained on ImageNet with CSQ. Specifically, BadNets is the most classic backdoor poison-label backdoor attack by modifying the pixel point of the image as a trigger. To get a better attack performance, we choose a square position with size of 18 in the lower right corner to change the pixel point to white and set poisoning rate to be 5\%. 
We use PGD~\cite{kurakin2018adversarial} to add noise to the image and then paste a distinct patch of size 20 in the bottom right corner as a trigger.
We set the poisoning rate in line with our attack.
ISSBA is the latest invisible backdoor attack that writes noise into an image as a trigger through image watermarking techniques, but need to modify the label. We set the poison rate of ISSBA to be 5\%. 
We add all of the poisoned samples into the corresponding training set.

\noindent\textbf{Analysis.} The comparison results of different attack methods are shown in Table \ref{tab:Comparison Experiment}. For t-MAP our attack performs best. Specifically, compared to BadNets we achieve invisible trigger with clean label, compared to LCBA we hide the trigger, and compared to ISSBA we do not need to modify the labels of the poisoned samples. Our poisoning ratio is lower than  BadNets and ISSBA. The targeted retrieval results on ImageNet and MS-COCO are shown in Figure~\ref{Fig.pr}, which also show our attack outperforms all the other methods.

As shown in  Figure~\ref{fig:comparation}, the visual stealthiness of the images produced by various methods is consistent with   our quantitative evaluation results. Our method produces images with higher MSE and PSNR than all other methods, and higher SSIM than LSBA and ISSBA. BadNets slightly outperforms BadHash with regard to SSIM metrics because it produces images with small local changes. In the first row of Figure~\ref{fig:comparation}, changing pixel values as triggers is easily to be recognized for BadNets. 
As shown from the third row of  Figure~\ref{fig:comparation}, even though the watermark added by ISSBA is well concealed, it can be easily distinguished when added to  images with strong chromatic aberrations, and the watermark in the residual map is  conspicuous. 


\vspace{-0.2cm}\section{conclusion}\label{conclusion}

In this paper, we propose the first truly imperceptible backdoor attack against deep hashing that fulfills the requirements of visual invisibility and clean label setting simultaneously. 
We present a new cGAN network combined with a label-based contrastive learning network to generate invisible and input-specific poisoned images with clean label.
Our extensive experiments on benchmark datasets ImageNet and MS-COCO verify that BadHash is highly effective at attacking state-of-the-art deep hashing schemes. The experiment results also demonstrate that the poisoned samples generated by BadHash have desirable invisibility and transferability.

\section*{Acknowledgments}
Shengshan’s work is supported in part by the National Natural Science Foundation of China (Grant No. U20A20177). 
Yifeng’s work is supported in part by the Guangdong Basic and Applied Basic Research Foundation (Grant No. 2021A1515110027). 
Yuanyuan’s work is supported in part by the National Natural Science Foundation of China (Grant No. 62002127). Yuanyuan He is the corresponding author.



\bibliographystyle{ACM-Reference-Format}
\balance
\footnotesize
\bibliography{acmart}
\end{document}